\documentclass[nonacm,sigconf]{acmart}
\usepackage{tabularx}
\usepackage{multirow}
\usepackage{booktabs}
\usepackage{balance}
\usepackage{appendix}

\begin{document}

\title{Benefit from Reference: Retrieval-Augmented Cross-modal Point Cloud Completion}

\author{Hongye Hou}
\orcid{0009-0000-2850-3425}
\affiliation{%
  \institution{Xi'an Jiaotong University}
  \city{Xi'an}
  \country{China}
}
\email{houhongye2001@stu.xjtu.edu.cn}

\author{Zhan Liu}
\orcid{0009-0001-2062-7914}
\affiliation{%
  \institution{Xi'an Jiaotong University}
  \city{Xi'an}
  \country{China}
}
\email{predator@stu.xjtu.edu.cn}

\author{Yang Yang}
\authornote{Corresponding author}
\orcid{0000-0001-8687-4427}
\affiliation{%
  \institution{Xi'an Jiaotong University}
  \city{Xi'an}
  \country{China}
}
\email{yyang@mail.xjtu.edu.cn}

\renewcommand{\shortauthors}{Someone et al.}

\begin{abstract}
Completing the whole 3D structure based on an incomplete point cloud is a challenging task, particularly when the residual point cloud lacks typical structural characteristics. Recent methods based on cross-modal learning attempt to introduce instance images to aid the structure feature learning. However, they still focus on each particular input class, limiting their generation abilities. In this work, we propose a novel retrieval-augmented point cloud completion framework. The core idea is to incorporate cross-modal retrieval into completion task to learn structural prior information from similar reference samples. Specifically, we design a Structural Shared Feature Encoder (SSFE) to jointly extract cross-modal features and reconstruct reference features as priors. Benefiting from a dual-channel control gate in the encoder, relevant structural features in the reference sample are enhanced and irrelevant information interference is suppressed. In addition, we propose a Progressive Retrieval-Augmented Generator (PRAG) that employs a hierarchical feature fusion mechanism to integrate reference prior information with input features from global to local. Through extensive evaluations on multiple datasets and real-world scenes, our method shows its effectiveness in generating fine-grained point clouds, as well as its generalization capability in handling sparse data and unseen categories.

\end{abstract}
 
%   <concept_id>10010147.10010371.10010396</concept_id>
%   <concept_desc>Computing methodologies~Shape modeling</concept_desc>
%   </concept>

% \ccsdesc[500]{Computing methodologies~Shape modeling}

\keywords{Point Cloud Completion; Generative Model; 3D-Retrieval}

	\maketitle
	\section{Introduction}
    \begin{figure}[htbp]
		\centering
		\includegraphics[scale=0.38]{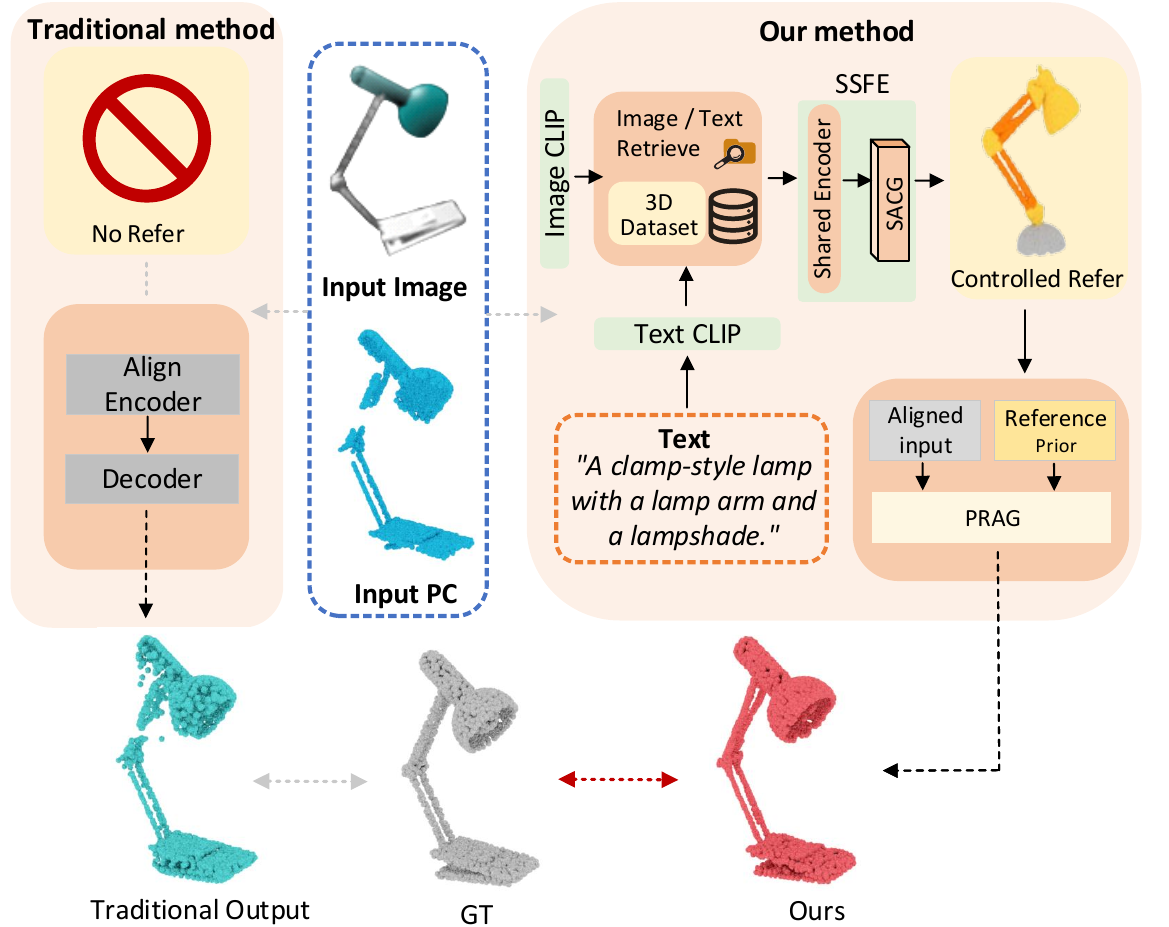}
		\caption{Compared with the traditional method and our method under the encoder-decoder framework. The main difference is that cross-modal (text or image) retrieval is introduced into the point cloud completion. More structural prior information from similar reference samples can be utilized to generate missing parts jointly.}
		\label{fig:Itroduction}
        \vspace{-0.4cm}
	\end{figure}
    With the development of 3D computer vision, point cloud data is increasingly applied in various fields, such as embodied intelligence \cite{James2021CoarsetoFineQE}, automatic driving \cite{Chen20203DPC} and 3D scene understanding \cite{8954028}. However, due to the inherent limitations of scanning conditions, like viewing angle occlusions and surface reflectivity, the raw point cloud data often exhibit incompleteness. Recovering complete and high-fidelity 3D point clouds is crucial for many downstream tasks \cite{Liang2018DeepCF,Nie2020RfDNetPS}.
    
    Deep Neural Networks have been widely and successfully used for 3D point cloud feature encoding \cite{charlesPointNetDeepLearning2017, wangDynamicGraphCNN2019, zhaoPointTransformer2021, 10658198}. In this case, current methods for point cloud completion are usually formulated in an encoder-decoder framework \cite{yuanPCNPointCompletion2018, huangPFNetPointFractal2020} as shown in Fig. \ref{fig:Itroduction}, which learn latent structural patterns from incomplete inputs and generate complete object in 3D space. Although these approaches have achieved promising results, they suffer from two potential limitations: (1) Structural Generalization Limitation: the structure feature relies on a data-driven training manner. When real-world data contain arbitrary rotation angles, unseen category or sparser presentations, the feature may not be generated based on limited input information. (2) Loss of Detail Information: for detail-rich targets, it is extremely challenging to infer details of missing structures from partial inputs. Therefore, some methods \cite{zhang2021vipc, aiello2022crossmodallearningimageguidedpoint} introduce instance images captured by RGB cameras on 3D scanners to guide generation. However, the inherent differences between different modalities impact the effectiveness for generating fine-grained details. \par
Recall that when a human attempts to repair an unseen structure, his brain first imagines a similar object that has been seen before. Useful reference structures are then filtered out to help integrate with the original structure. Thus, instead of focusing on inputs, we propose to incorporate cross-modal (text or image) retrieval into point cloud completion framework and reformulate the completion task as a joint generation problem, based on cross-modal inputs and 3D reference sample as shown in Fig. \ref{fig:Itroduction}. In this idea, there are two additional requirements for the completion networks: (1) the encoder should identify and learn on relevant structural features from reference samples, (2) the decoder should effectively leverage original inputs and reference prior features.\par

With the development of multi-modal pre-trained neural network models like Contrastive Language-Image Pre-Training (CLIP) \cite{radford2021learningtransferablevisualmodels}, similar samples can be easily searched from the prepared cross-modal database according to the input image or text description. To achieve the above joint generation goal, we propose a Retrieval-Augmented cross-modal point cloud completion framework. As shown in Fig. \ref{fig:Itroduction}, we design a Structural Shared Feature Encoder (SSFE) with a core component called Similarity \& Absence Control Gates (SACG). SACG firstly calculates the similarity of structural features within the input context and identifies the intersection between reference and input features. Then, one similarity control gate learns relevant structural features. The other absence control gate suppresses irrelevant information interference. Finally, reference features are reconstructed to obtain structural priors useful for the missing parts. In the decoding stage, we propose a Progressive Retrieval-Augmented Generator (PRAG), that fuses reference and input features. Specifically, PRAG employs a global-to-local cross-attention mechanism to promote the interaction generation. Their global information is merged via pooling to construct an initial seed. Subsequently, component-level attentional interactions guided by semantic information enable the transfer of local geometric details. Based on these two components, the point cloud completion network learns more structural prior information from similar reference samples to generate rich geometric details for the missing part. Finally, our method has the generalization capability in handling sparse data and unseen categories. 

	The main contributions of this work are summarized as follows: 
	\begin{itemize}
		\item {We propose a novel retrieval-augmented point cloud completion framework inspired by the human brain's structural repair reasoning. By incorporating cross-modal retrieval, our method gains additional structural priors to generate missing parts, achieving state-of-the-art performance on multiple benchmarks and real-world scenes.}
		\item {We design the SSFE encoder as an effective adaptive feature extraction module to jointly extract cross-modal features and reconstruct reference features. Benefiting from the SACG mechanism, relevant structural features are enhanced and irrelevant information interference is suppressed.}
		\item {We also design the PRAG decoder that employs a hierarchical feature fusion to integrate reference structural priors with input features. From global to local levels, PRAG guides the quality of completion and further enriches geometric details.}
	\end{itemize}
	
	\section{Related Work}
	\subsection{3D Shape Generation}
	In recent years, significant progress has been made in 3D shape generation methods driven by various inputs, including interaction modes such as text \cite{zhang2024claycontrollablelargescalegenerative}, images \cite{huang2025spar3dstablepointawarereconstruction}, and incomplete point clouds \cite{yuanPCNPointCompletion2018}. We focus on shape generation using 3D point clouds as representations. Early generation models based on deep learning mainly used voxel-based representations \cite{wangOCNNOctreebasedConvolutional2017} to accomplish geometric inference by predicting voxel occupancy. However, these models are limited by the computational cost, which increases significantly with resolution and the blurring of surface details. PointNet \cite{charlesPointNetDeepLearning2017} accomplished geometric inference by predicting voxel occupancy. PCN \cite{yuanPCNPointCompletion2018} was the first end-to-end point cloud completion framework, pioneering the classical architecture of encoding into global variable-feature decoding. Subsequent research attempted to generate more fine-grained point clouds. SeedFormer \cite{zhouSeedFormerPatchSeeds2022} proposed seed-based staged generation steps that employed an up-sampling transformer to incrementally generate missing structures. SnowflakeNet \cite{Xiang2021SnowflakeNetPC} simulated the generation process of point clouds as the real-world growth of snowflakes, proposing a Snowflake Point De-convolution strategy and introducing a novel jump transformer to learn the splitting patterns. AdaPoinTr \cite{yuAdaPoinTrDiversePoint2023} employed Transformer \cite{vaswaniAttentionAllYou2017} to map input local point proxies to seed point proxies and utilized local geometric relations to recover detailed geometric structures. However, these methods are limited by incomplete inputs and perform poorly in the face of sparser and unseen category.
	
	\subsection{Multi-modal Point Cloud Completion}
   Directly predicting the missing structures from local point clouds is a challenging task in point cloud completion. To address these issues, Key-prompt \cite{2024houMM} alleviated information loss by using semantic associations to identify and learn similar structures from the input point cloud. ShapeNet-ViPC \cite{zhang2021vipc} introduced image information, utilizing a modality converter to transform images directly into skeletal point clouds, which were then combined with occluded point clouds. XMFnet \cite{aiello2022crossmodallearningimageguidedpoint} reduced the discrepancy between image and point cloud features and employed cross-attention mechanisms to fuse them. EGIInet \cite{xu2024explicitlyguidedinformationinteraction} trained both 2D and 3D encoders simultaneously and aligned modalities directly during training, ensuring interaction between missing image features and point cloud features while minimizing information loss. During the modality alignment, information loss can hinder accurate reconstruction of missing structures from images. SDFusion \cite{cheng2023sdfusionmultimodal3dshape} reconstructed complete 3D point clouds by combining monocular images and incomplete inputs, encoding point cloud priors into intermediate representations such as SDF \cite{chen2019learningimplicitfieldsgenerative}, and using diffusion models as decoders for 3D reconstruction. However, these methods often result in a loss of geometric details during feature interaction, hindering the achievement of high-fidelity, fine-grained reconstructions.
   	\begin{figure*}[htbp]
		\begin{center}
			\includegraphics[scale=0.47]{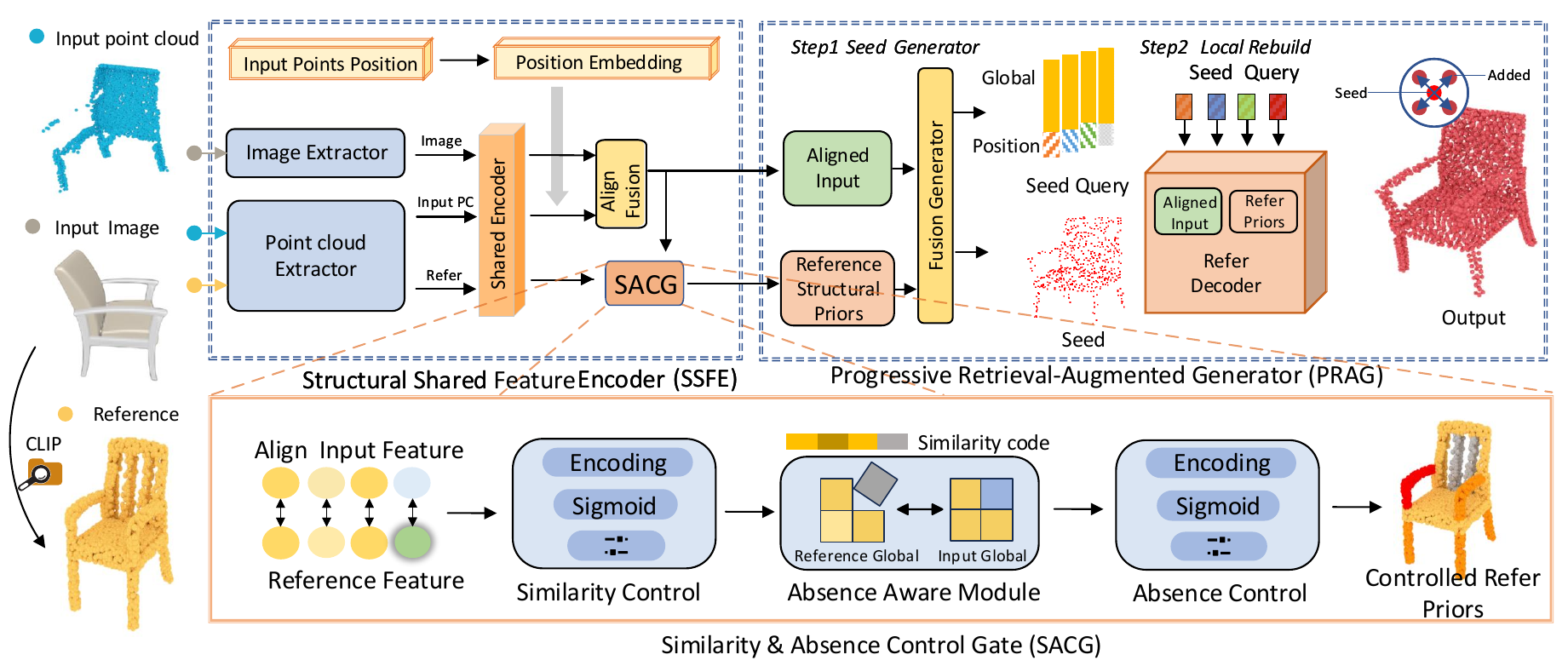}
			
		\end{center}
		\caption{Overview of the proposed retrieval-augmented point cloud completion framework. Given an incomplete 3D point cloud and its image, we first retrieve one similar point cloud as reference from a 3D dataset. In the encoding stage, the SSFE extracts structure features for both input and reference samples. Especially in the encoding process, SACG is proposed to reconstruct the prior information of reference structures, reduce noise, and enhance similar structures. In the decoding stage, the PRAG integrates features to generate complete point clouds with geometric details from global to local.}
		
		\label{fig:pipeline2}
	\end{figure*}
    \subsection{Retrieval-Augmented Generation}
    Previous Retrieval-Augmented Generation (RAG) aimed to improve language \cite{Retrieval-augmented} and image generation \cite{Ashual2022KNNDiffusionIG} by incorporating relevant external information during the generation process. While traditional point cloud completion methods also attempted to provide geometric information for missing regions using a dataset of 3D shapes. For example, researchers at Stanford University used nonrigid alignment of context models \cite{Example-based3D} with input data through warping techniques. However, these methods are encumbered by high inference optimization and database construction costs. They are also significantly sensitive to noise. Recently, Phidias \cite{wang2025phidias} introduced retrieval models to 3D Artificial Intelligence Generated Content, which used meta-control diffusion networks and routing modules to manage reference models across various similarity levels. However, diffusion-based information fusion reduces the fidelity of generated content and requires rotating the reference model. In contrast, we propose a retrieval-augmented point cloud completion approach, which extracts valuable geometric priors while maximizing the use of input information. Our method is able to ensure high-fidelity 3D reconstruction without complex view rotations or extensive databases.

	\section{Method}
	\subsection{Method Overview}
Given an incomplete 3D point cloud $P \in \mathbb{R}^{N \times 3}$, the cross-modal completion task is to recover its 3D structure with the help of the single-view image $I \in \mathbb{R}^{H \times W \times C}$. Inspired by the structural-repair reasoning process of the human brain, our main idea is to refer to similar 3D objects and use relevant structure features as prior information to generate the missing part. Based on this, a novel retrieval-augmented point cloud completion method is proposed. 
Figure \ref{fig:pipeline2} exhibits the overview of our framework. A cross-modal dataset is pre-built by expanding the 3D point cloud dataset with rendered images. Based on the multi-modal pre-trained neural network model CLIP \cite{radford2021learningtransferablevisualmodels}, a similar reference sample can be easily searched according to Image or Text Encoder (see Section 4.1.1 for more details). Then, the completion task is reformulated as a joint generation problem based on cross-modal inputs and the 3D reference sample. The architecture of our method consists of two key components: (1) Structural Shared Feature Encoder (SSFE), an effective adaptive feature extraction module using proposed Similarity \& Absence Control Gates (SACG) to promote feature interaction across reference and input data. Benefiting from the dual-channel control gates, relevant structural features are enhanced and irrelevant information interference is suppressed. (2) Progressive Retrieval-Augmented Generator (PRAG), a hierarchical feature fusion module to integrate reference structural priors with input features. From global to local levels, PRAG guides the quality of complete point cloud, and further enriches geometric details.

	\subsection{ Structural Shared Feature Encoder}
	
	%\subsubsection{Structural Shared Feature Encoder}
    % To effectively capture the local structural relationships among input point cloud, image, and reference 3D sample, we design a shared structural encoder. 
    For misaligned cross-modal inputs and reference data, each point and image patch is represented by \textit{local proxies} according to the structural information of their K-Nearest neighbors. Different from the commonly used serialized encoding techniques like EGIINet \cite{xu2024explicitlyguidedinformationinteraction}, local proxies notice the localized structure and long-range interactions. Our encoder also avoids the absolute positional encoding, effectively mitigates spatial misalignment due to pose changes of the reference samples. 
    %After the shared encoding stage, we would fuse aligned features of cross-modal inputs, which helps to remain the global structural features contained in the input image. 
    \par
    
    Especially, for the image \( I \), we employ a patch-based encoding technique, dividing it into a certain number of regions, which are then transformed into feature vectors \( \mathbf{F}_i \) via 2D convolution. 
\begin{equation} 
		\mathbf{F}_{i} = \mathrm{Conv2D}(\mathrm{Patch}(I))
\end{equation}
For the input point cloud \( P \) and the reference  point cloud \( P_r \), we utilize a regional proxy encoding method, where a single point aggregates its neighborhood to represent the relative structural relationships within the neighborhood. This idea of using aggregated local features to a single point is shown to be applicable in point cloud feature extraction \cite{10658198}. We also use ball query to identify neighboring points. Compared to K-Neighbor search, it better captures the structural information of key structures. As shown in the Equation~\eqref{eq:quadratic}, the aggregated relative positions are subsequently encoded using graph convolution.
\begin{equation} 
		\mathbf{F}_{p} = \mathrm{GraphConv}(\mathbf{F}_{p}-\text{BallQuery}(P_i,\mathbf{F}_{p})) \label{eq:quadratic} 
\end{equation}
\par \textbf{Shared Encoder:} To effectively capture the local structural relationships among image, input point cloud, and reference 3D sample, we design a shared structure encoder. By leveraging the self-attention mechanism in Vision-Transformer \cite{dosovitskiy2021imageworth16x16words} modules, our model effectively captures crucial long-range unified information among different modalities and different objects within the same space. Notably, positional encoding is applied only to the input point cloud. Unlike traditional approaches, it omits absolute position embedding for retrieved point cloud features $\mathbf{F}_{p'}$, which omission enables the model to learn structural information from retrieval point clouds, regardless of the point clouds' poses.
\begin{equation}
    \mathcal{F}_{I}, \mathcal{F}_{p}, \mathcal{F}_{p'} = 
    \mathrm{SFE}(\mathbf{F}_{i}), 
    \mathrm{SFE}\left( \mathbf{F}_{p} + \mathrm{Pos}(P) \right),
    \mathrm{SFE}(\mathbf{F}_{p'})
\end{equation}
After the shared encoding, we fuse aligned features of cross-modal inputs, which helps retain the global structural features contained in the input image. This helps the model to learn the global information of missing structures from images. 
For the reference point cloud, this encoder avoids interference from absolute positional discrepancies, facilitating long-range interactions and preventing misalignment issues in subsequent processes. \par

\textbf{Similarity \& Absence Control Gates (SACG)} : To effectively focus on information related to similarities and absent parts from the reference sample, we propose the dual-channel control gate called Similarity \& Absence Control Gates (SACG). The first gate is used to encode feature relevance, masking out irrelevant components in the reference samples while enhancing the impact of relevant parts. The second gate is designed to sense absent components. We combine similar features with the global input point cloud for encoding and amplify the influence of missing components. The simultaneous use of these two gates allows us to obtain beneficial information from various reference samples.\par
    \begin{figure}[htbp]
		\centering
		\includegraphics[scale=0.4]{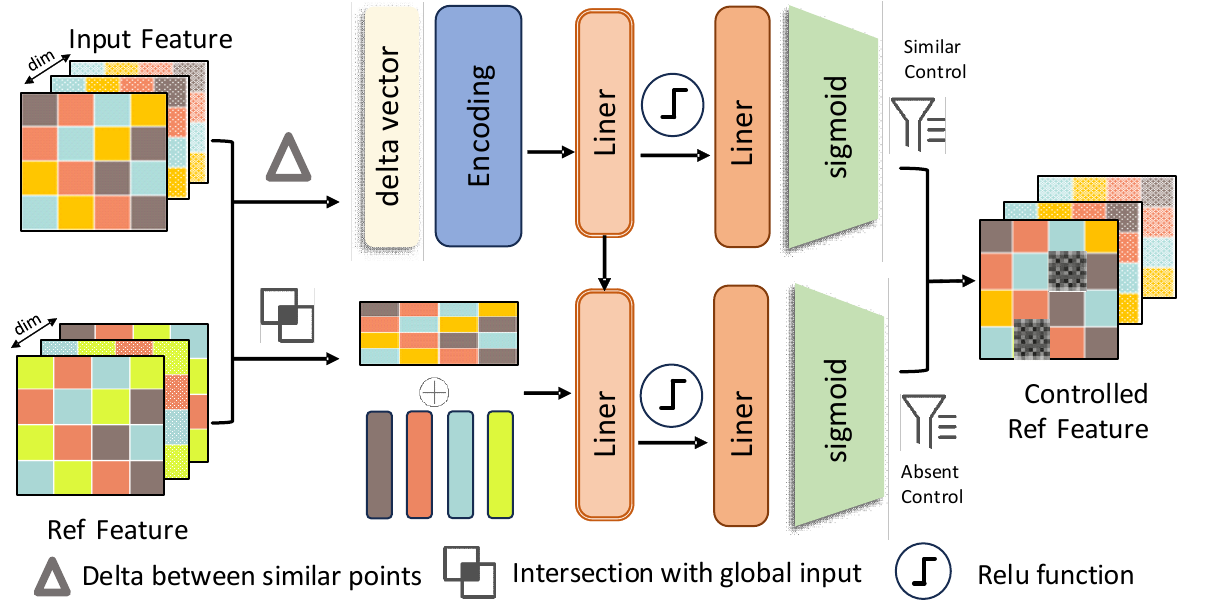}
		\caption{ The network structure of the SACG. It encodes differences in feature similarity and the intersection of input structural features. The sigmoid function is used to control the output. Thereby the corresponding features are filtered or enhanced during the feature reconstruction.}
		\label{fig:method_SACG}
	\end{figure}
Specifically, we show the network structure of SACG in Fig. \ref{fig:method_SACG}. The two gates are computed using delta and intersection. Relying on SSFE, we extract features from the relative positional relationships within the neighborhood, which are rich in semantic information due to the fact that similar structural parts have similar relative positional characteristics. For each point's feature $F_{p'_i}$ ,we locate the four most similar neighbors in $F_{p}$ based on semantic similarity, and calculate the feature difference between the point and its neighbors as a similarity delta encoding. This encoding is processed through a multilayer perceptron (MLP) and transformed into a similarity gate using the sigmoid function. As shown in Equation~\eqref{eq:quadratic2}, $\kappa$ represents the nearest neighbor features are found based on feature similarity, and $\sigma$ is the sigmoid function.
\begin{equation} 
    S_{i} = \sigma(\mathrm{MLP}(\mathcal{F}_{p{\prime}}^{i} - \mathcal{F}_{p}^{l})), \forall l: \mathcal{F}_{p^{l}} \in \kappa(\mathcal{F}_{p{\prime}}^{i}) \label{eq:quadratic2}
\end{equation}
We extract the global features $G_{p}$ of the input by increasing the dimension and taking the maximum over the rows. Next, we concatenate each $F_{p'_i}$ with $G_{p}$ and combine it with the similarity encoding of that point, then encode the result using an MLP. This process helps determine whether a point in the reference sample is located in a missing or critical focus area of the input point cloud, such as a missing or incomplete boundary.

\begin{equation}
    \mathcal{C}_{p'_i} = \mathrm{MLP}(S_{i} \cdot (\mathcal{F}_{p'}^{i} \oplus G_{p}))
\end{equation}

Here, the dimension of the gating matrix $\mathcal{C}_{p'}$ is $\mathbb{R}^{N \times \text{dim}}$, N is the number of reference points proxies.
	
 	\subsection{Progressive Retrieval-Augmented Generator}
In this section, we present a novel module called Progressive Retrieval-Augmented Generator (PRAG) for decoding stage. The generation process of PRAG leverages the reconstructed structured-encoded retrieval features as auxiliary tools to infer missing parts based on the existing shape structure and recover geometric details while maintaining data fidelity. Due to the effective handling of reference sample by the control gates, we propose a progressive assistance scheme to benefit from it. Initially, a complete yet sparse point cloud, referred to as the "seed," is generated by coupling the global variables of the input point cloud and the reference samples. Using the seed as an intermediate variable, we further learn details from both the input and retrieval models to decode the local neighborhood structure of the seed. During this step-wise generation process, we progressively learn global to local levels knowledge from the aligned input and the reference features processed by the control gates.

Specifically, with the help of the SSFE module, we achieve modality alignment between images and point clouds, and perform interactive fusion of their structural information. We aim to realize cross-domain feature interaction between the input and retrieval point clouds in three-dimensional space during generation. The retrieval point cloud effectively provides geometric priors for the missing structures in the input, leading PRAG to first employ a fusion generator that combines input information and retrieval priors’ global knowledge to generate a seed representing the overall contour.

	\begin{equation} 
		p_{q}^{i}= MLP[\mathbf{G}] =MLP[Max(\mathcal{F}_{p'_i}),Max(\mathcal{C}_{p'_i} \cdot \mathcal{F}_{p'_i}) ]
	\end{equation}
To restore the local details of the seed, we aim to re-represent the seed features to reflect its local neighborhood information. Thus, we first generate the local query of the seed using global variables and positional information:
\begin{equation} 
		\mathcal{Q}_{i}= MLP[\mathbf{G},p_{q}^{i}] 
	\end{equation}

Previous decoder architectures typically rely on cross-attention mechanisms to learn relevant information from the input. However, since only a subset of the components in the retrieval model is relevant to the input, cross-modal feature interaction becomes particularly important. Thanks to the previously adopted structural encoding, the semantics of the reference model are aligned with those of the input point cloud. Therefore, through semantic relevance, we can search the most relevant point clouds, allowing the model to focus on similar structures in the retrieval model proxy $\mathcal{F}_{p{\prime}}$ during decoding. 
	\begin{figure}[htbp]
		\centering
		\includegraphics[scale=0.39]{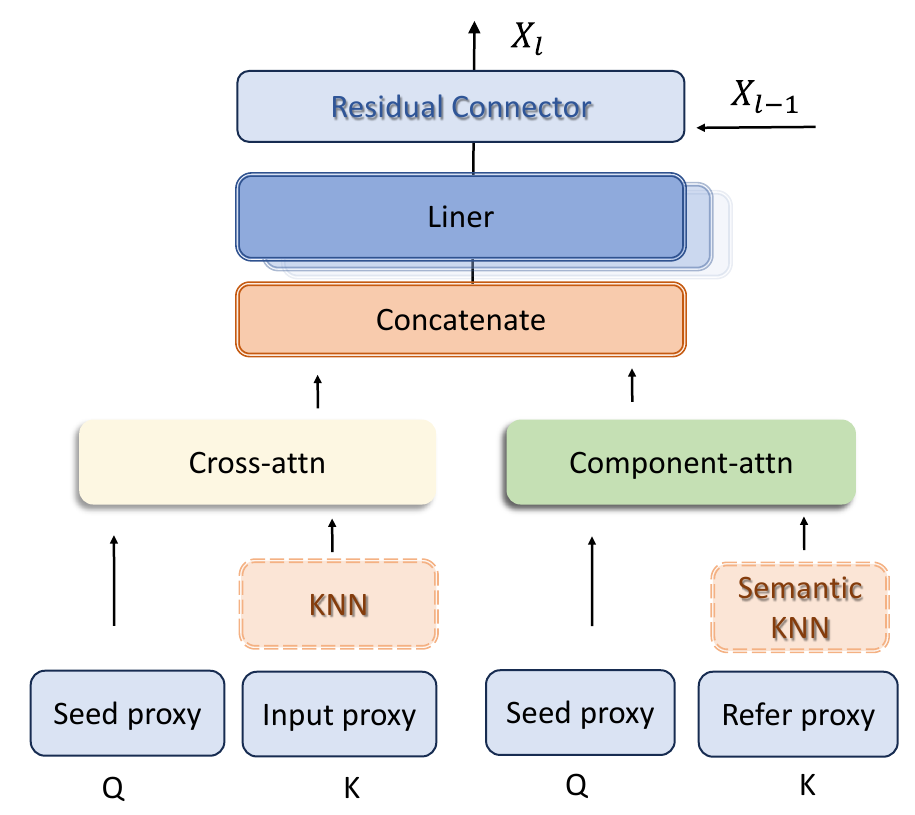}
		\caption{ Architecture of refer decoder. For input and reference features, we have taken geometric KNN search and semantic KNN as part of the Transformer block respectively.}
		\label{fig:Method2}
	\end{figure}
As illustrated in the Fig. \ref{fig:Method2}, in the specific implementation of local structure decoding, we first identify several retrieval proxies with similar features in the retrieval model through semantic similarity to represent the most similar components. We then apply local component attention mechanisms for learning:
	\begin{equation} 
		\tau \left(\mathcal{Q}_{i}\right)=\operatorname{Cross-attn}(\mathcal{Q}_{i}, \mathcal{F}_{p{\prime}}^{l}-\mathcal{Q}_{i}), \forall l: \mathcal{F}_{p{\prime}}^{l} \in \kappa\left(Q_{i}\right),
	\end{equation}
Subsequently, a simple MLP module is used to convert the seed proxy \(Q\) into displacement shifts \(H\) for neighboring points, refining the sparse seed into a dense and complete point cloud. Finally, we obtain a point cloud \(Z \subseteq \mathbb{R}^{M \times 3}\) composed of \(M\) points:

	\begin{equation} 
		\mathbf{Z}_{i}^{k} = \mathcal{H}_{i}^{k} + p_{q}^{i}, k = \frac{M}{M_0}
	\end{equation}
	where $M_0$ denotes the number of seed points, and $k$ represents the number of localized points per seed. This results in a point cloud $\mathbf{Y} \subseteq \mathbb{R}^{M \times 3}$ containing $M = M_0 \times k$ points.
	
	\subsection{Loss Function}
    The loss function for point cloud completion should be a good geometric quantitative measure of the output quality. The most commonly used is Chamfer Distance (CD) \cite{Fan2016APS}, which calculates the Euclidean distance of each point from its nearest neighbor found in the target space, which is an O(N log N ) complexity algorithm.
    \begin{equation} 
		D_{\mathrm{CD}}\left(P_{1}, P_{2}\right)=\frac{1}{P_{1}} \sum \min _{y \in P_{2}}\|x-y\|_{2}^{2}+\frac{1}{P_{2}} \sum \min _{x \in P_{1}}\|y-x\|_{2}^{2}
	\end{equation}
    Since we use a hierarchical generation approach, we first down-sample the truth value to 512 points to compute $\mathcal{L}_{seed}$ is used to constrain the seed generation process. In order to evaluate the quality of the final refined generation results, comparison with the ground truth produces a loss of final results denoted as $\mathcal{L}_{output}$.
    \begin{equation}
		\mathcal{L}_{seed} = D_{\mathrm{CD}}(p_{q},\mathbf{Y}_{gt}^{1}), \mathcal{L}_{output} = D_{\mathrm{CD}}(p_{q}, \mathbf{Y}_{gt})
	\end{equation}
    \par A very important task in multi-modal point cloud completion is to align the image features with the point cloud features. In addition to the interaction based on the direct cross-attention, there are also methods that design a supervised approach Feature Transfer-loss \cite{xu2024explicitlyguidedinformationinteraction}, which realizes the interaction of key structural information in image features and point cloud features by calculating the MSE of the GRAM matrices of the two features, and at the same time, FT-loss also constrains the 3D features of the point cloud before and after encoding to avoid the structural changes during the interaction.
\begin{equation}
\mathcal{L}_{\text{FT}} = \frac{\sum \left(\boldsymbol{G}(\boldsymbol{F}_{in}) - \boldsymbol{G}(\boldsymbol{F}_{out})\right)^2}{N \times Dim} +(\boldsymbol{F}_{in} - \boldsymbol{F}_{out})^2
\end{equation}
	As illustrated in the equation, we denote "in" and "out" to represent the inputs and outputs of the SSFE for images and point clouds, respectively. $\boldsymbol{G}$ represents the GRAM matrix computed for the features. It is crucial to note that the modalities of input and output need to be crossed to fully exploit the complementary information between images and point clouds. For the interaction between images and point clouds, mutual calculations and summations are required. Additionally, we perform an extra computation for the gated reference and input point clouds, which will supervise the enhanced SACG ability to retain more relevant components.
    \par The final loss consists of three parts: $\mathcal{L}_{\text{seed}}$ for seeding the multi-stage reconstruction, $\mathcal{L}_{\text{output}}$ for the deviation of the final output from the true value, and $\mathcal{L}_{\text{FT}}$ for feature alignment and interaction.
	\begin{equation}
		\mathcal{L} = \mathcal{L}_{seed}+\mathcal{L}_{output}+\mathcal{L}_{FT}
	\end{equation}

	\section{EXPERIMENTS}
	\begin{table*}
\caption{Completion results on ShapeNet-ViPC dataset in terms of per-point L2 Chamfer Distance ×1000 (lower is better) and F1-score.}
\label{tab:VIPC_results}
\begin{tabular}{c|c|cccccccc|cc}
\toprule
Category & Method & Plane & Cabinet & Car & Chair & Lamp & Couch & Table & Boat & Avg (CD-$\ell_1$) & Avg (F-Score@1\%) \\
\midrule
\multirow{8}{*}{Only 3D Input}
& FoldingNet \cite{yangFoldingNetPointCloud2018} & 5.242 & 6.958 & 5.307 & 8.823 & 6.504 & 6.368 & 7.080 & 3.882 & 6.271 & 0.331 \\
& AtlasNet \cite{Groueix2018APA} & 5.032 & 6.414 & 4.868 & 8.161 & 7.182 & 6.023 & 6.561 & 4.261 & 6.062 & 0.410 \\
& PCN \cite{yuanPCNPointCompletion2018} & 4.246 & 6.409 & 4.840 & 7.441 & 6.331 & 5.668 & 6.508 & 3.510 & 5.619 & 0.407 \\
& TopNet \cite{tchapmiTopNetStructuralPoint2019} & 3.710 & 5.629 & 4.530 & 6.391 & 5.547 & 5.281 & 5.381 & 3.350 & 4.976 & 0.467 \\
& PF-Net \cite{huangPFNetPointFractal2020} & 2.515 & 4.453 & 3.602 & 4.478 & 5.185 & 4.113 & 3.838 & 2.871 & 3.873 & 0.551 \\
& GRNet \cite{Xie2020GRNetGR} & 1.916 & 4.468 & 3.915 & 3.402 & 3.034 & 3.872 & 3.071 & 2.160 & 3.171 & 0.601 \\
& PoinTr \cite{yuPoinTrDiversePoint2021} & 1.686 & 4.001 & 3.203 & 3.111 & 2.928 & 3.507 & 2.845 & 1.737 & 2.851 & 0.683\\
& SeedFormer \cite{zhouSeedFormerPatchSeeds2022} & 1.716 & 4.049 & 3.392 & 3.151 & 3.226 & 3.603 & 2.803 & 1.679 & 2.902 & 0.688 \\
& AdaPoinTr \cite{yuAdaPoinTrDiversePoint2023} & 1.716 & 4.049 & 3.392 & 3.151 & 3.226 & 3.603 & 2.803 & 1.679 & 2.423 & 0.705 \\
% & Our  & 1.716 & 4.049 & 3.392 & 3.151 & 3.226 & 3.603 & 2.803 & 1.679 & 2.902 & 0.688 \\
\hline
\multirow{4}{*}{With 2D Guided}
& VIPC \cite{zhang2021vipc} & 1.760 & 4.558 & 3.183 & 2.476 & 2.867 & 4.481 & 4.990 & 2.197 & 3.308 & 0.591 \\
& CSDN \cite{10015045} & 1.251 & 3.670 & 2.977 & 2.835 & 2.554 & 3.240 & 2.575 & 1.742 & 2.570& 0.695 \\
& XMFNet \cite{aiello2022crossmodallearningimageguidedpoint} & 0.572 & 1.980 & 1.754 & 1.403 & 1.810 & 1.702 & 1.386 & 0.945 & 1.443 & 0.796 \\
& EGIINet\cite{xu2024explicitlyguidedinformationinteraction} & 0.534 & 1.921 & 1.655 & 1.204 & 0.776 & 1.552 & 1.227 & 0.802 & 1.211 & 0.836 \\
& Ours & \textbf{0.517} & \textbf{1.626} & \textbf{1.537} & \textbf{1.057} & \textbf{0.643} & \textbf{1.126} & \textbf{1.097} & \textbf{0.673} & \textbf{0.988} & \textbf{0.889} \\
\bottomrule
\end{tabular}
\end{table*}

	In this section, we conduct extensive experiments to validate the superiority of our method. We evaluate our approach on the ShapeNet-ViPC dataset \cite{zhang2021vipc}, including its unseen categories and a sparser variant with noisy inputs. Furthermore, we perform experiments on the KITTI dataset \cite{geigerAreWeReady2012}, which consists of RGB images and sparse point clouds captured from real-world scenes. Both the quantitative metrics and visual results of our method demonstrate superior performance.
	
	\subsection{Implementation Details}
	\subsubsection{Retrieval and Setting Details}In order to obtain a reference point cloud, we construct a 3D model dataset based on the ShapeNet dataset and objaverse dataset \cite{deitke2022objaverseuniverseannotated3d} with their rendered 12 images of each object. In use, the corresponding models can be retrieved by image CLIP \cite{radford2021learningtransferablevisualmodels} embedding or text. When it is not feasible to retrieve using rendered images, we also encode the dataset using ULIP \cite{xue2023uliplearningunifiedrepresentation}, and retrieve using the encoding of incomplete point clouds. In addition, users can obtain reference point clouds by generating 3D models from pictures or text \cite{huang2025spar3dstablepointawarereconstruction,xiang2024structured3dlatentsscalable,zhang2024claycontrollablelargescalegenerative,wang2025phidias}. We utilize two NVIDIA A100 GPUs and employed Adam \cite{kingma2017adammethodstochasticoptimization} as the optimizer, setting the initial learning rate to $$2 \times 10^{-4}$$, and 160 epochs will be conducted with a learning rate decay set to 0.7. The ablation study is conducted under the same experimental conditions.
	
	\subsubsection{Evaluation Metrics} 
    To quantify the completion performance, as in previous work, we use Chamfer Distance (CD) \cite{Fan2016APS} and F-score \cite{Tatarchenko2019WhatDS} as quantitative evaluation metrics. Specifically, CD increases significantly when the generated point cloud contains extra or missing parts compared to the ground truth, and a lower value of this is better; F-score is used to evaluate the proportion of similar components, and a higher value is better.

	\begin{figure*}[htbp]
		\begin{center}
			
			\includegraphics[scale=0.32]{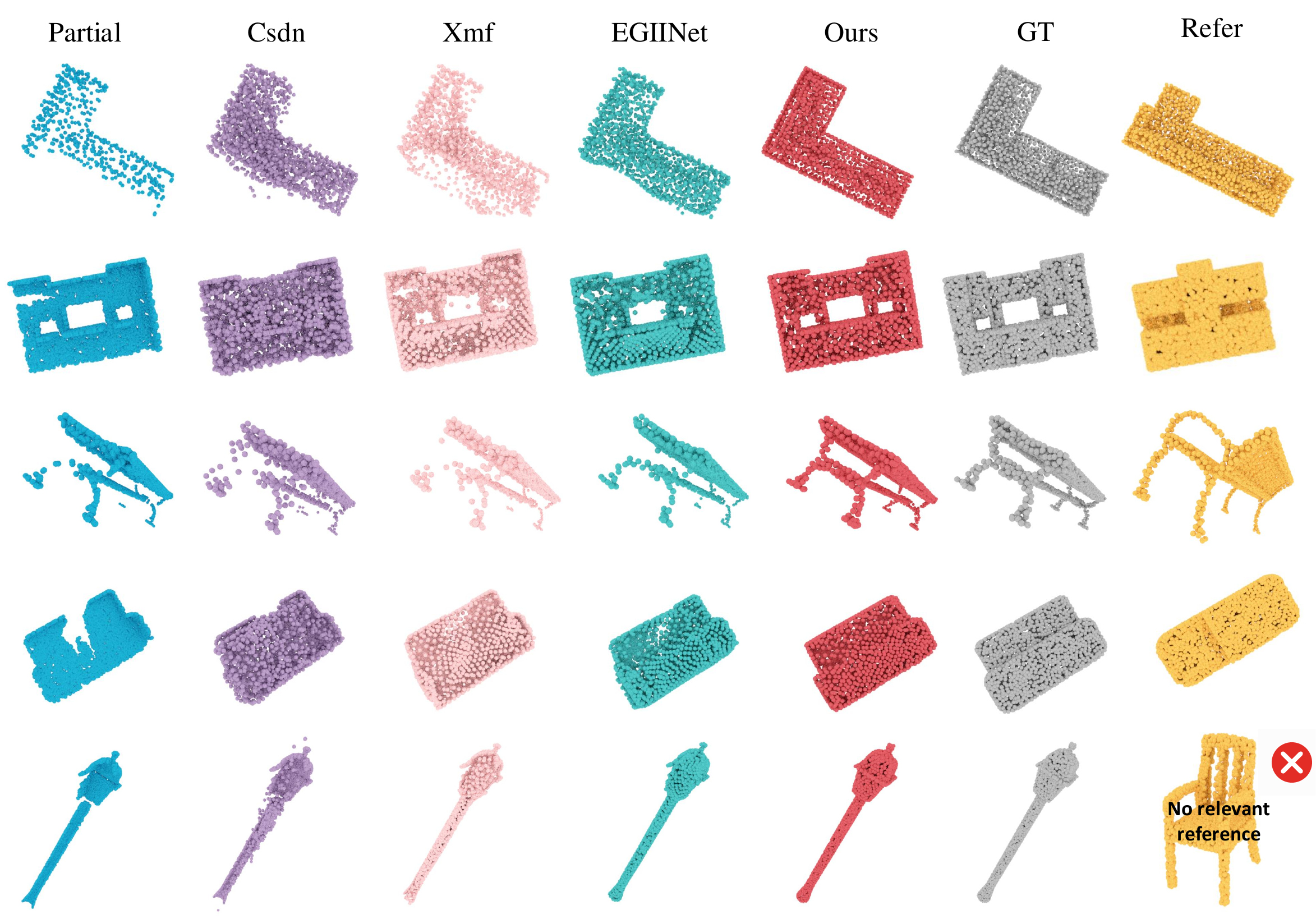}
			
		\end{center}
		
		\caption{Qualitative comparisons on the ShapeNet-ViPC dataset.}
		\label{fig:vis results}
	\end{figure*}
	\subsection{Multi-modal Point Cloud Completion}
	\subsubsection{Evaluation on ShapeNet-ViPC Dataset and Unseen Categories}
\textbf{Data.} The ShapeNet-ViPC dataset \cite{zhang2021vipc} comprises 38,328 objects spanning 13 categories. Each object in this dataset has a missing point cloud constructed from 24 viewpoints, with the same viewpoint setup as ShapeNetRendering  \cite{xu2019disndeepimplicitsurface}. Unlike the PCN dataset \cite{yuanPCNPointCompletion2018} and ShapNet-55/34 datasets \cite{yuAdaPoinTrDiversePoint2023}, each 3D shape is rotated to match the pose corresponding to a specific viewpoint, allowing for a broader range of rotation angles. In our experiments, we adhere to the dataset setup described in ViPC \cite{zhang2021vipc} for both training and testing to ensure comparability with existing models. To evaluate the generalization ability and robustness of the model, we pre-trained the model on 8 categories in the ShapeNet-ViPC dataset and evaluated it on the remaining 5 unseen categories, including monitor and speaker, which are not part of the training set and other categories.

\textbf{Results.} In Tab. \ref{tab:VIPC_results}, we compare the performance of our proposed method with current models such as PoinTr \cite{yuPoinTrDiversePoint2021} and AdaPoinTr \cite{yuAdaPoinTrDiversePoint2023} in the case of 3D-only inputs and current advanced methods under multi-modal inputs scenarios. Our approach achieved superior performance across all categories, showing improvements of up to 0.2 reduction in CD and 5\% enhancements in F1 scores. Furthermore, we illustrate the qualitative results for selected categories, which demonstrate the integration of retrieval-based prior knowledge significantly enhances the generation of detailed structures. We present a visual comparison of our method with previous approaches in Fig. \ref{fig:vis results}, where it can be intuitively observed that our method achieves more realistic and accurate results compared to prior methods. Benefiting from our improved encoder, the proposed method maintains good performance even when no relevant reference is found.\par
\textbf{Results on Unseen Scenes.} For the evaluation on unseen categories, our method demonstrates notable advancements and robust generalization capabilities, as shown in Tab. \ref{tab:unseen_results}. By leveraging retrieval-augmented networks, our approach effectively captures precise prior information from references, enabling it to perform well on categories not encountered during training. These results highlight the method's ability to generalize and produce accurate outcomes even for previously unseen data. \textit{More visualization results can be found in the Appendix.}
    \begin{table}[!htbp]
	
	\caption{Completion results on ShapeNet-VIPC Unseen dataset}
	\label{tab:unseen_results}
	\small
	\begin{tabular}{p{2.8cm}@{\hspace{-2em}}|ccccc}
		
		\toprule
		\multicolumn{1}{c}{\textbf{}} & \multicolumn{5}{c}{\textbf{5 unseen categories}} \\
		\cmidrule(lr){2-6}
		& \footnotesize Bench & \footnotesize Monitor & \footnotesize Speaker & \footnotesize CD-$\ell_1$ & \multicolumn{1}{c}{\footnotesize F-Score} \\
		\midrule
		PF-Net \cite{yuanPCNPointCompletion2018} & 3.683 & 5.304 & 7.663 & 5.011& 0.468 \\
		MSN \cite{tchapmiTopNetStructuralPoint2019} & 2.613 & 4.818 & 8.259 & 4.684 & 0.533 \\
		GRNet \cite{Xie2020GRNetGR} & 2.367 & 4.102 & 6.493 & 4.096 & 0.548 \\
		PoinTr \cite{yuPoinTrDiversePoint2021} & 1.976 & 4.084 & 5.913 & 3.755 & 0.619 \\
		PointAttN \cite{wang2022pointattnneedattentionpoint} & 2.135 & 3.741 & 5.973 & 3.674 & 0.605 \\
        SDT \cite{9804851} & 4.096 & 6.222 & 9.499 & 6.001 & 0.327 \\
		\midrule
        VIPC\cite{zhang2021vipc} &3.091 & 4.419 &7.674 & 4.601 & 0.498 \\
        CSDN\cite{10015045} &1.834& 4.115 & 5.690 & 3.656 & 0.631 \\
        XMFNet\cite{aiello2022crossmodallearningimageguidedpoint} & 1.278 & 2.806 & 4.823 & 2.671 & 0.710 \\
        EGIINET\cite{xu2024explicitlyguidedinformationinteraction} & 1.047 & 2.513 & 4.282 & 2.354 & 0.750 \\
		Ours & \textbf{0.923} & \textbf{1.743} & \textbf{3.591} & \textbf{1.834} & \textbf{0.822} \\
		\bottomrule
	\end{tabular}
\end{table}
\begin{table*}
    % \footnotesize
    \caption{KITTI Dataset results. The comparison between the following models is based on the FD and MMD metrics.}

    \label{tab:KITTI}
    \begin{tabular}{c|ccccccc|cc}
        \toprule
        \footnotesize CDl2(x1000) & \footnotesize AtlasNet \cite{Groueix2018APA} & \footnotesize PCN \cite{yuanPCNPointCompletion2018} & \footnotesize PFNet \cite{huangPFNetPointFractal2020} & \footnotesize GRNet \cite{Xie2020GRNetGR} & \footnotesize SeedFormer \cite{zhouSeedFormerPatchSeeds2022} & \footnotesize PoinTr \cite{yuPoinTrDiversePoint2021} & \footnotesize AdaPoinTr \cite{yuAdaPoinTrDiversePoint2023} & \footnotesize EGIINet\cite{xu2024explicitlyguidedinformationinteraction} & \footnotesize \textbf{Ours} \\
        \midrule

        Fidelity &1.879 &2.435 &1.247 &0.916 &0.311 &\textbf{0} &0.337 &\textbf{0} &0.116 \\
        MMD &2.308 &1.566 &0.992 &0.972 &0.716 & 0.709 &0.522 &0.516 &\textbf{0.281} \\
        \bottomrule
    \end{tabular}

\end{table*}
\begin{table*}
\caption{Completion results on sparse and noisy scene}
\label{tab:sparse_VIPC_results}
\begin{tabular}{c|cccccccc|cc|c}
\toprule
Method & Plane & Cabinet & Car & Chair & Lamp & Couch & Table & Boat & Avg (CD-$\ell_1$) & Avg (F-Score@1\%) & Avg Reduction \\
\midrule
% VIPC \cite{zhang2021vipc} & 1.760 & 4.558 & 3.183 & 2.476 & 2.867 & 4.481 & 4.990 & 2.197 & 3.308 & 0.591 & $2.47 \downarrow$ \\
CSDN \cite{10015045} & 1.790 & 4.542 & 3.568 & 3.965 & 4.319 & 4.092 & 3.644 & 2.666 & 2.570->3.573  & 0.695->0.513 & $0.182 \downarrow$ \\
XMFNet \cite{aiello2022crossmodallearningimageguidedpoint} & 1.506 & 3.841 & 3.175 & 2.919 & 2.265 & 3.958 & 2.923 & 1.727 & 1.443->2.789 & 0.796->0.608 & $0.188 \downarrow$ \\
EGIINet \cite{xu2024explicitlyguidedinformationinteraction} & 1.285 & 3.840 & 2.897 & 2.456 & 1.790 & 2.839 & 2.494 & 1.384 & 1.211->2.373 & 0.836->0.669 & $0.167 \downarrow$ \\
Ours & \textbf{0.780} & \textbf{2.110} & \textbf{1.993} & \textbf{1.428} & \textbf{0.968} & \textbf{1.473} & \textbf{1.777} & \textbf{0.968} & \textbf{0.988->1.434} & \textbf{0.889->0.818} & \textbf{0.071} $\downarrow$ \\
\bottomrule
\end{tabular}
\end{table*}

	\subsection{Completion on Real Scenes }
	\subsubsection{Evaluation on KITTI}
	\par \textbf{Data.} To evaluate our model's performance with real-world data, we conduct experiments on the KITTI \cite{geigerAreWeReady2012} dataset, which is sourced from LIDAR scans. Recognized widely in autonomous driving research, the KITTI dataset presents challenges due to the sparsity inherent in LIDAR-derived data. So, generating complete and dense point clouds is essential for downstream tasks like 3D target detection. Since this dataset does not provide complete point clouds as ground truth, we follow the approach of GRNet \cite{Xie2020GRNetGR}, using Fidelity Distance (FD) and Minimal Matching Distance (MMD) as evaluation metrics. In addition, we reconstructed a extend dataset containing image input patterns based on labels for 2D and 3D target detection also containing the category pedestrians.
	
\begin{figure}[htbp]
		\centering
		\includegraphics[scale=0.22]{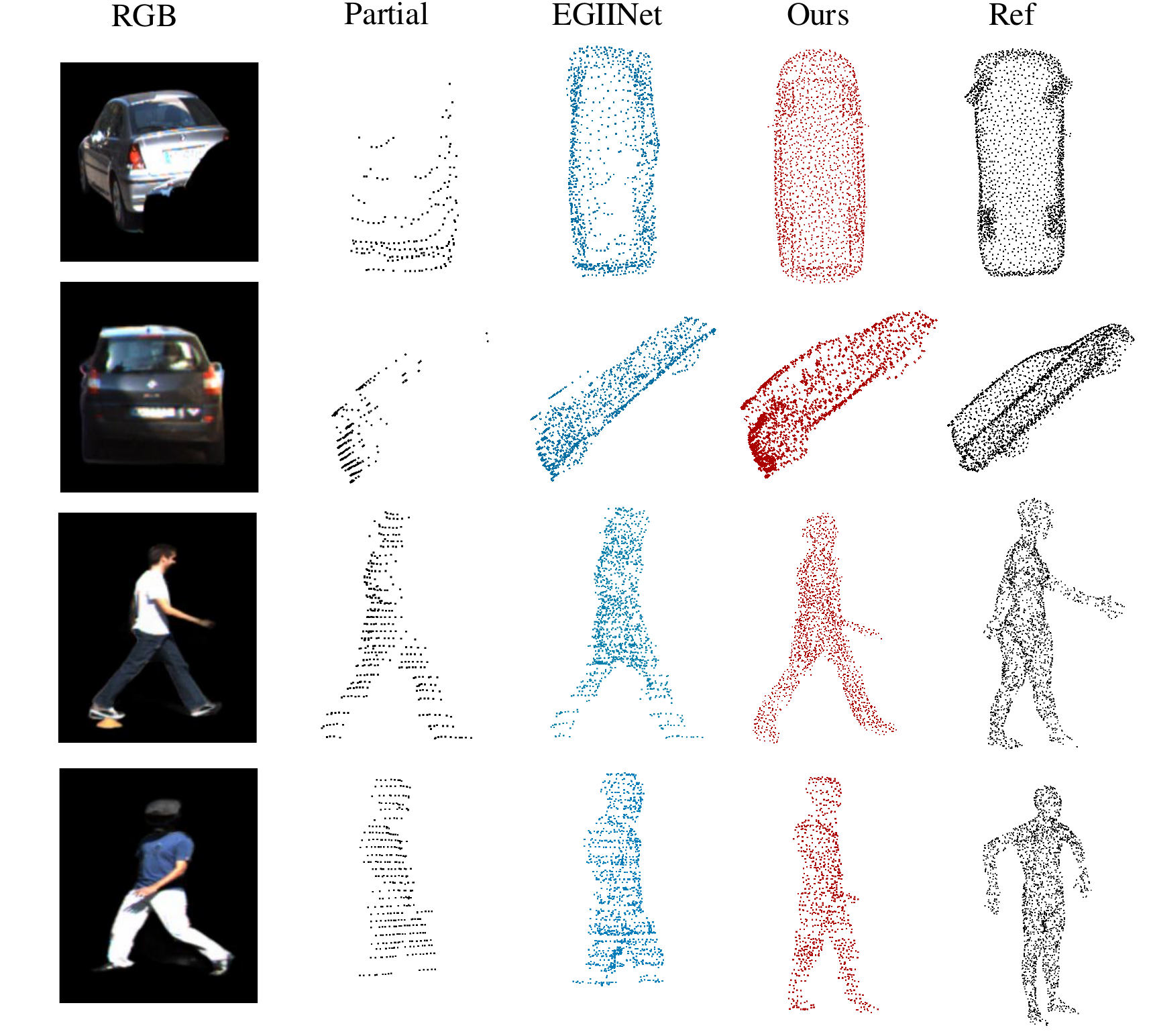}
		\caption{ Qualitative results on the KITTI dataset. We show two different views of each object while our method can recover a car with more accurate contour and details.}
		\label{fig:KITTI results}
	\end{figure}

	\textbf{Results on Real Scenes.}
	We initially trained our model using the ShapeNetViPC-Dataset \cite{yuanPCNPointCompletion2018} to supplement the incomplete car and pedestrian data in KITTI. For previous methods with single-modal inputs, we conducted training on the PCN dataset following the GRNet approach. As demonstrated in Tab. \ref{tab:KITTI}, our model surpasses several baseline models in performance. Since PoinTr and EGIINet splice the inputs into the final result, the value of FD is 0, but this is not robust in the face of noise. As shown in Fig. \ref{fig:KITTI results}, vehicle point clouds generated by our method contains high-fidelity details, such as front and rear mirrors. As an unseen category in model training, other methods produce poor results for pedestrians. However, our method is still able to fill in the missing arms and lower limbs of pedestrians.

\subsubsection{Evaluation on Sparse and Noisy Scenes.}
    \par \textbf{Data.} In real-world scenes, point clouds obtained from LiDAR are often sparse and noisy. Therefore, we simulate this scene by constructing a more difficult emulation dataset based on the existing benchmark \cite{zhang2021vipc} to test the model's ability to handle sparse and noisy incomplete point clouds. In the experimental setup, we reduce the input from the original 2048 points to 256 points. Additionally, we introduce noise into the input point cloud following the noise construction method in AdaPoinTr \cite{yuAdaPoinTrDiversePoint2023}, using random Gaussian distribution. The ground truth data remains at 2048 points, requiring the generation of an equal number of points under sparser conditions.\par
	
	\textbf{Results on sparse and noisy scenes.}
	For the performance comparison, we evaluate the CD, as well as the degradation of the metric values relative to the standard input conditions. Tab. \ref{tab:sparse_VIPC_results} gives statistical results for eight different categories of more challenging sparse and noisy point clouds. Due to the introduction of the reference prior information, our method shows no significant performance degradation and exhibits excellent results compared to other methods.

	\subsection{Ablation Study}
    To validate the effectiveness of each module design, we conduct a series of detailed ablation experiments on the key modules in Tab. \ref{tab:AblationStudyImproved}. Specifically, we analyze the shared encoder proposed for the encoding stage and the gating mechanism used to process retrieved point clouds, as well as different approaches for utilizing reference samples in the decoding stage. Additionally, we consider the special case where the retrieval input is irrelevant.\par
\textbf{Introducing Retrieval Priors}: We build a 3D dataset and introduced retrieval prior information to assist point cloud generation. XMFNet \cite{aiello2022crossmodallearningimageguidedpoint} is used as the baseline for the ablation comparison. The experimental results of model A, introducing retrieval priors indeed leads to some improvement in performance. However, differences of pose and structural between the retrieved reference and input limit the generation quality, resulting in less significant metrics improvement.\par

\textbf{Shared Encoder and Positional Encoding}: This module addresses the spatial misalignment between the retrieved 3D models and the input point cloud, as well as the gap between the incomplete and complete point cloud structures. The effectiveness of the shared module is validated by Model B in the table, where the use of a shared encoder reduces the Chamfer Distance by 0.04. Model B1 also uses the shared encoder and adds positional encoding. The comparison between Model B and B1 demonstrates that the absolute position encoding of the input point cloud, commonly adopted in previous methods, has a negative impact, hindering the interaction of long-range structural information.\par

\textbf{Similarity \& Absence Control Gates (SACG)}: This module aims to filter and process irrelevant feature parts in the reference samples. We conduct two sets of experiments: Model C utilizes the control gates to handle similar retrieval objects. The results show that using SACG to extract features from input point clouds significantly reduced CD to 1.06. Experiment Model C1 still uses the control gates but assumes that completely irrelevant objects are retrieved. In this case, the model degrades to the level of Method B, which does not cause significant negative impact.\par

\textbf{Progressive Retrieval-Augmented Generator (PRAG)}: We test the generation ability during decoding. Experiment Model D adopts a step-by-step seed generation approach but still uses global cross-attention in phase two. The final method integrates our all innovative modules, achieving state-of-the-art results.\par
    \begin{table}[!htbp]
    \footnotesize
    \centering
    \caption{Ablation Study. The table proves the validity of our three module designs respectively.}
    \label{tab:AblationStudyImproved}
    \begin{tabular}{@{}c|c|c|c|cc@{}}
        \toprule
        % 添加的新行
        & \multicolumn{2}{c|}{Encoding Stage} & Decoding Stage & & \\
        \cmidrule(lr){2-3} \cmidrule(lr){4-4}
        Model & Align Encoder & Refer Process & Retrieve-Enhanced & CD & F1 \\
        \midrule
        XMFNet & Cross-attn & - &  - & 1.443 & 0.796 \\
        A & Cross-attn & - & Cross-attn & 1.361 & 0.811 \\
        B & Shared ViT & - & Cross-attn & 1.314 & 0.830 \\
        B1 & Add Position & - & Cross-attn & 1.354 & 0.822 \\
        C & Shared ViT & SACG & Cross-attn & 1.144 & 0.845 \\
        C1 & Shared ViT & SACG & Cross-attn & 1.255 & 0.831 \\
        D & Shared ViT & SACG & 2 stage & 1.062 & 0.850 \\
        Ours & Shared ViT & SACG & PRAG & \textbf{0.988} & \textbf{0.889} \\
        \bottomrule
    \end{tabular}
\end{table}
    
	\section{Conclusion}
	
	In this paper, we presented an innovative and effective cross-modal point cloud completion framework assisted by 3D retrieval. Our method utilizes retrieved similar features as a priori knowledge to generate detailed missing structures. To achieve this goal, we design the structural encoder, which reconstructs retrieval features to ensure that the model can learn benefit structures from various retrieved point clouds. Additionally, we propose a progressive decoder that employs hierarchical feature fusion from global to local levels, which facilitating precise and gradual integrate between retrieval priors and input features. Experimental results demonstrate our outstanding performance on benchmark datasets and real-world scenes. In the future, we will try to introduce more multi-modal features to enrich this retrieval completion framework.
	\clearpage
    \bibliographystyle{ACM-Reference-Format}
    \balance
	\bibliography{keyprompt2.bib}

\end{document}